\documentclass{article}
\usepackage{spconf,amsmath,graphicx}
\usepackage{color}

\title{A 2D dilated residual U-Net for multi-organ segmentation in thoracic CT}
\name{Sulaiman Vesal, Nishant Ravikumar, Andreas Maier\thanks{Thanks to EFI-Erlangen for funding.}}
\address{Pattern Recognition Lab, Friedrich-Alexander-University Erlangen-Nuremberg, Germany}
\begin{document}
%
\maketitle
\begin{abstract}
Automatic segmentation of organs-at-risk (OAR) in computed tomography (CT) is an essential part of planning effective treatment strategies to combat lung and esophageal cancer. Accurate segmentation of organs surrounding tumours helps account for the variation in position and morphology inherent across patients, thereby facilitating adaptive and computer-assisted radiotherapy. Although manual delineation of OARs is still highly prevalent, it is prone to errors due to complex variations in the shape and position of organs across patients, and low soft tissue contrast between neighbouring organs in CT images. Recently, deep convolutional neural networks (CNNs) have gained tremendous traction and achieved state-of-the-art results in medical image segmentation. In this paper, we propose a deep learning framework to segment OARs in thoracic CT images, specifically for the: heart, esophagus, trachea and aorta. Our approach employs dilated convolutions and aggregated residual connections in the bottleneck of a U-Net styled network, which incorporates global context and dense information. Our method achieved an overall Dice score of 91.57\% on 20 unseen test samples from the ISBI 2019 SegTHOR challenge.  
\end{abstract}


\section{Introduction}
\label{sec:intro}
Organs at risk (OAR) refer to structures surrounding tumours, at risk of damage during radiotherapy treatment \cite{doi:10.1002/mp.12045}. Accurate segmentation of OARs is crucial for efficient planning of radiation therapy, a fundamental part of treating different types of cancer. However, manual segmentation of OARs in computed tomography (CT) images for structural analysis, is very time-consuming, susceptible to manual errors, and is subject to inter-rater differences\cite{doi:10.1002/mp.12045}\cite{7950685}. Soft tissue structures in CT images normally have very little contrast, particularly in the case of the esophagus. Consequently, an automatic approach to OAR segmentation is imperative for improved radiotherapy treatment planning, delivery and overall patient prognosis. 
Such a framework would also assist radiation oncologists in delineating OARs more accurately, consistently, and efficiently. Several studies have addressed automatic segmentation of OARs in CT images, with efforts being more focused on pelvic, head and neck areas \cite{doi:10.1002/mp.12045}\cite{7950685}\cite{Kazemifar_2018}. 

In this paper, we propose a fully automatic 2D segmentation approach for the esophagus, heart, aorta, and trachea, in CT images of patients diagnosed with lung cancer. Accurate multi-organ segmentation requires incorporation of both local and global information. Consequently, we modified the original 2D U-Net \cite{Unett}, using dilated convolutions \cite{Dilated} in the lowest layer of the encoder-branch, to extract features spanning a wider spatial range. Additionally, we added residual connections between convolution layers in the encoder branch of the network, to better incorporate multi-scale image information and ensure a smoother flow of gradients in the backward pass.

\begin{figure}[t!]
\begin{minipage}[b]{.48\linewidth}
  \centering
  \centerline{\includegraphics[width=4.0cm, height=3.5cm]{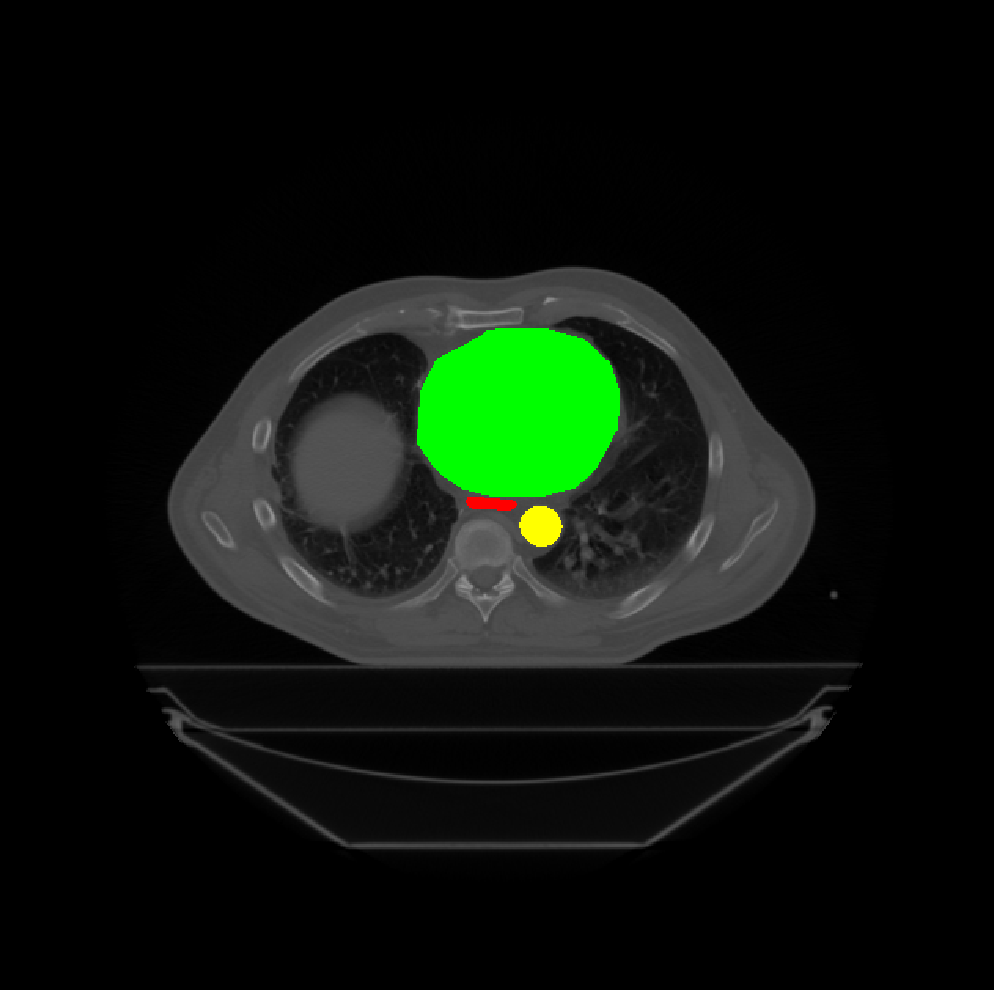}}

\end{minipage}
\hfill
\begin{minipage}[b]{0.48\linewidth}
  \centering
  \centerline{\includegraphics[width=4.0cm, height=3.5cm]{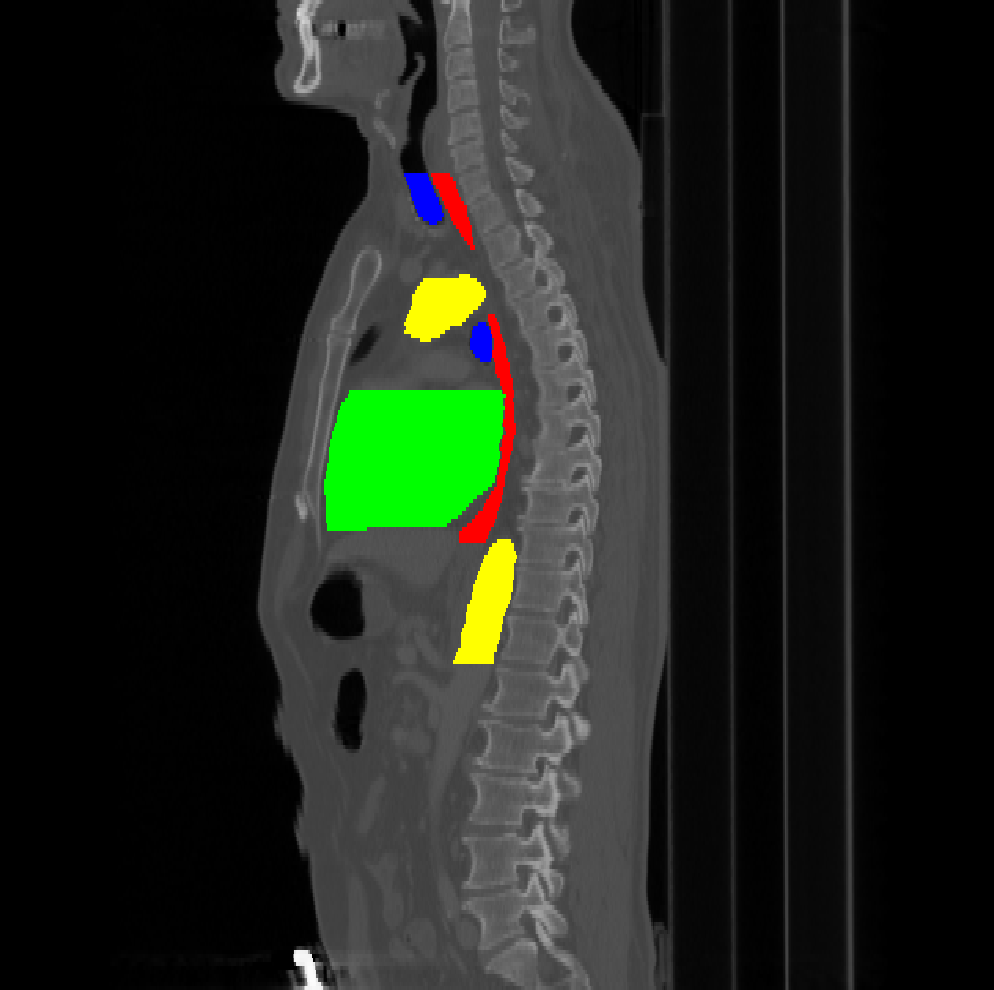}}
\end{minipage}
\begin{minipage}[b]{.48\linewidth}
  \centering
  \centerline{\includegraphics[width=4.0cm, height=3.5cm]{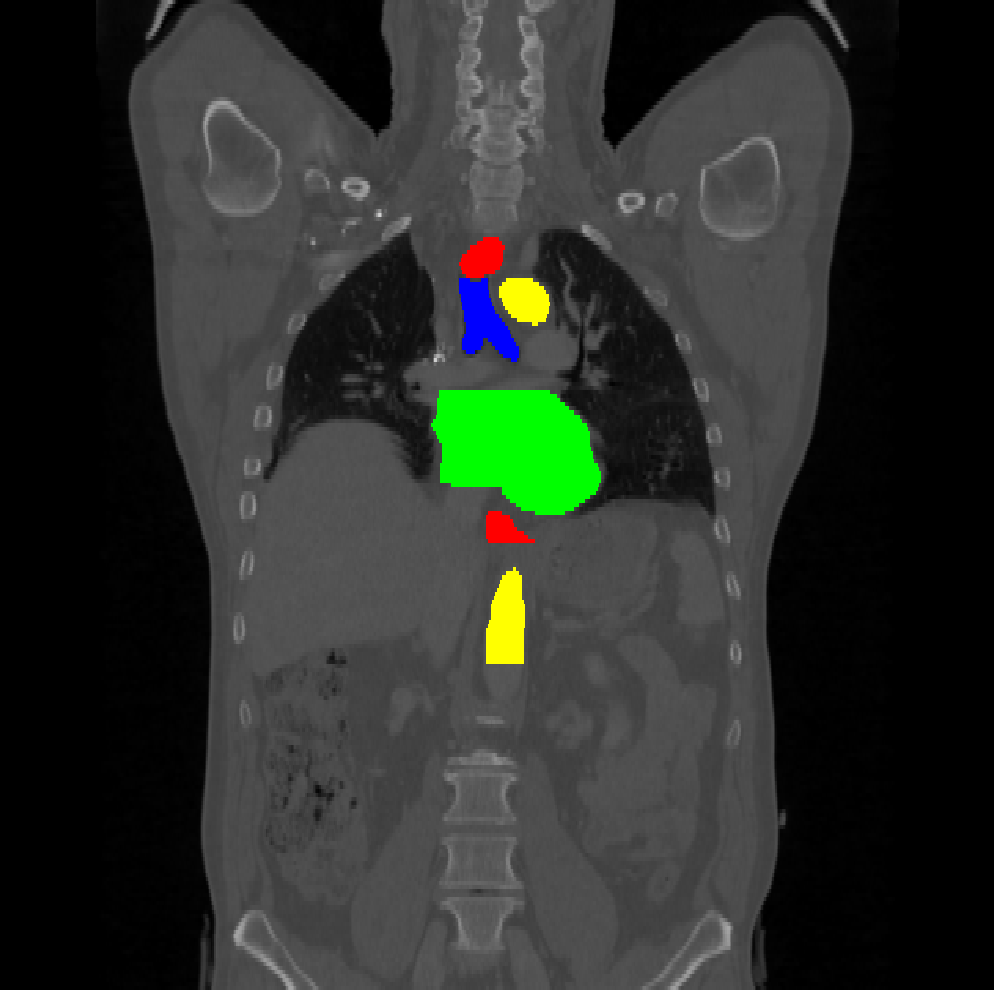}}
\end{minipage}
\hfill
\begin{minipage}[b]{0.48\linewidth}
  \centering
  \centerline{\includegraphics[width=4.0cm, height=3.5cm]{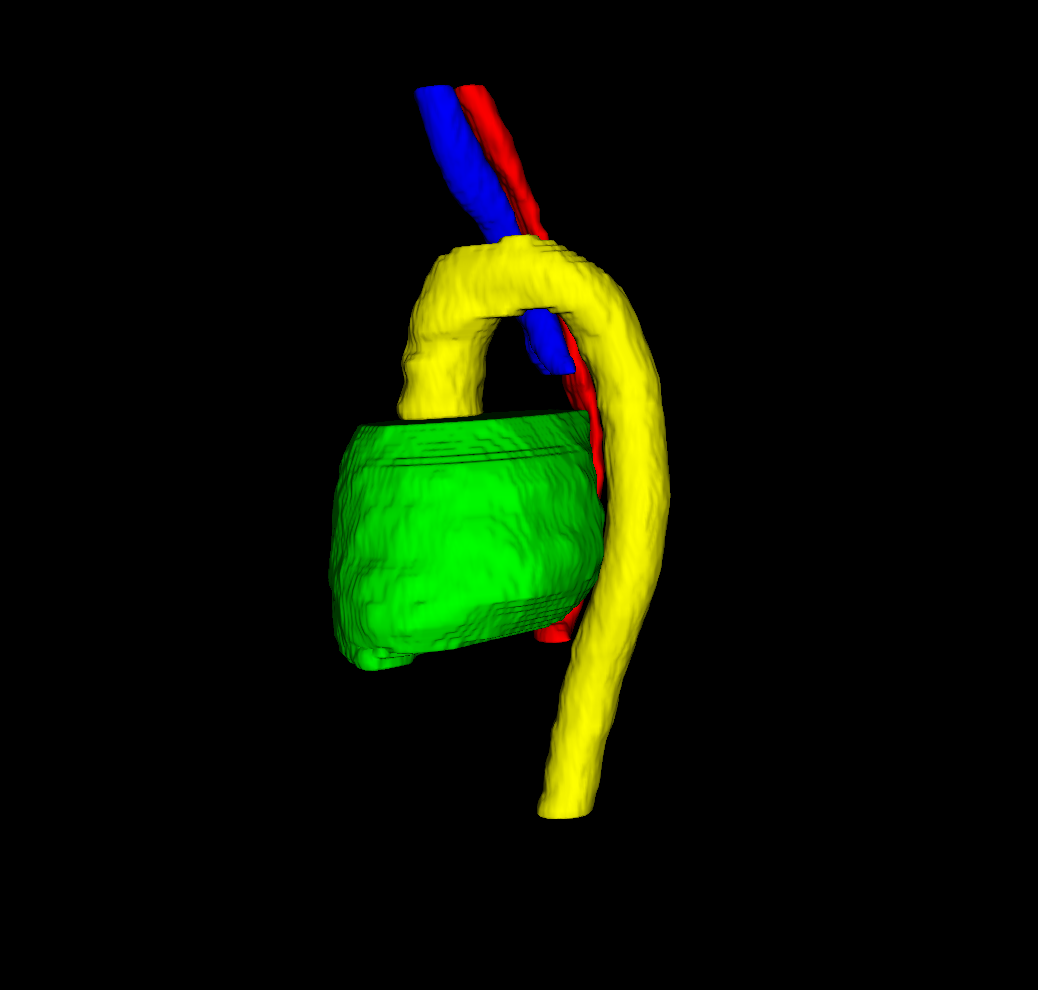}}
\end{minipage}
\caption{Example of OARs in CT images with axial, sagittal and coronal views and 3D surface mesh plot.}
\label{fig:res}
\end{figure}

\section{Methods}
Segmentation tasks generally benefit from incorporating local and global contextual information. In a conventional U-Net \cite{Unett} however, the lowest level of the network has a relatively small receptive field, which prevents the network from extracting features that capture non-local information. 
Hence, the network may lack the information necessary to recognize boundaries between adjacent organs, the fully connected nature of specific organs, among other properties that require greater global context to be included within the learning process. Dilated convolutions \cite{Dilated} provide a suitable solution to this problem. They introduce an additional parameter, i.e. the dilation rate, to convolution layers, which defines the spacing between weights in a kernel. This helps dilate the kernel such that a 3$\times$3 kernel with a dilation rate of $2$ results in a receptive field size equal to that of a 7$\times$7 kernel. Additionally, this is achieved without any increase in complexity, as the number of parameters associated with the kernel remains the same. 

\begin{figure}[htb]
  \centering
  \centerline{\includegraphics[width = 7.6 cm]{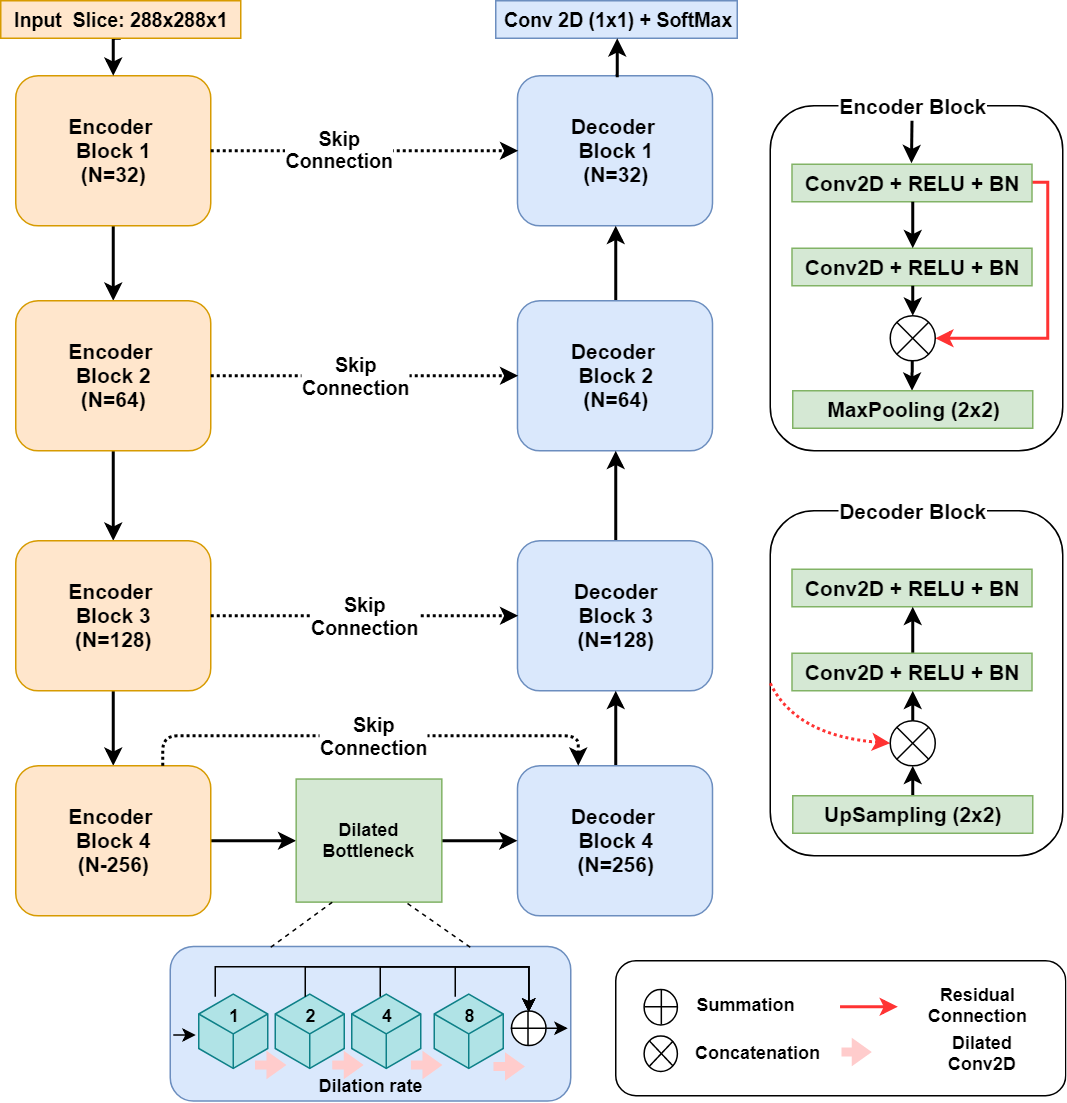}}
\caption{Block diagram of the 2D U-Net+DR architecture for thoracic OAR images segmentation. The left side shows the encoding part and the right side shows the decoding part. The network has four dilated convolutions in the bottleneck and residual connection in each encoder block respectively(shown in red color arrow).}
\label{fig:UNet}
\end{figure}

We propose a 2D U-Net+DR (refer to Fig.\ref{fig:UNet}.) network inspired by our previous studies \cite{SkinNet}\cite{DR_UNET}. It comprises four downsampling and upsampling convolution blocks within the encoder and decoder branches, respectively. In contrast to our previous approaches, here we employ a 2D version (rather than 3D) of the network with greater depth, because of the limited number of training samples. For each block, we use two convolutions with a kernel size of 3$\times$3 pixels, with batch normalization, rectified linear units (ReLUs) as activation functions, and a subsequent max pooling operation. Image dimensions are preserved between the encoder-decoder branches following convolutions, by zero-padding the estimated feature maps. This enabled corresponding feature maps to be concatenated between the branches. A softmax activation function was used in the last layer to produce five probability maps to distinguish the background from the foreground labels. Furthermore, to improve the flow of gradients in the backward pass of the network, the convolution layers in the encoder branch were replaced with residual convolution layers. In each encoder-convolution block, the input to the first convolution layer is concatenated with the output of second convolution layer (red line in Fig. \ref{fig:UNet}), and the subsequent 2D max-pooling layer reduces volume dimensions by half. The bottleneck between the branches employs four dilated convolutions, with dilation rates $1-4$. The outputs of each are summed up and provided as input to the decoder branch.

\subsection{Dataset and Materials}
\label{sec:format}
The ISBI SegTHOR challenge\footnote{https://competitions.codalab.org/competitions/21012} organizer provided the computed tomography (CT) images from the medical records of 60 patients. The CT scans are 512 $\times$ 512 pixels in size, with an in-plane resolution varying between 0.90 mm and 1.37 mm per pixel. The number of slices varies from 150 to 284 with a z-resolution between 2mm and 3.7mm. The most common resolution is 0.98$\times$0.98$\times$2.5 $mm^{3}$. The SegTHOR dataset (60 patients) was randomly split into a training set: 40 patients(7390 slices) and a testing set: 20 patients(3694 slices). The ground truth for OARs was delineated by an experienced radiation oncologist \cite{7950685}. 

\subsection{Pre-Processing}
Due to low-contrast in most of CT volumes in the SegTHOR dataset, we enhanced the contrast slice-by-slice, using contrast limited adaptive histogram equalization (CLAHE), and normalized each volume with respect to mean and standard deviation.  In order to retain just the region of interest (ROI), i.e. the body part and its anatomical structures, as the input to our network, each volume was center cropped to a size of 288$\times$288 along the $x$ and $y$ axes, while the same number of slices along $z$ were retained. We trained the model using the provided training samples via five-fold cross-validation (each fold comprising 32 subjects for training and 8 subjects for validation). Moreover, we applied off-line augmentation to increase the number of subjects within the training set, by flipping the volumes horizontally and vertically.  

\subsection{Loss Function}
In order to train our model, we formulated a modified version of soft-Dice loss \cite{VNet} for multiclass segmentation. Here the Dice loss for each class is first computed individually and then averaged over the number of classes. Let's suppose for the segmentation of an N$\times$N input image (CT slice with esophagus, heart, aorta, trachea and background as labels), the outputs are five probabilities with classes of $k= 0, 1, 2, 3, 4$, such that $\sum_{k}\hat{y}_{n,k} = 1$ for each pixel. Correspondingly, if ${y}_{n,k}$ is the one-hot encoded ground truth of that pixel, then the multiclass soft Dice loss is defined as follows:

\begin{equation}
\label{eq15}
	\zeta_{dc}(y, \hat{y})  = 1- \frac{1}{N}(\sum_{k}\frac{\sum_{n}y_{nk} \hat{y}_{nk}}{\sum_{n}y_{nk} + \sum_{n}\hat{y}_{nk}})
\end{equation}
In Eq. (\ref{eq15}) $\hat{y}_{nk}$ denotes the output of the model, where $n$ represents the pixels and $k$ denotes the classes. The ground truth labels are denoted by $y_{nk}$.

Furthermore, in the second stage of the training (described in detail in the next section), we used Tversky Loss (TL)\cite{Tversky}, as the multiclass Dice loss does not incorporate a weighting mechanism for classes with fewer pixels. 
The TL is defined as following:
\begin{equation}
\label{eq2}
   TL(y, \hat{y}) = 1- \frac{\sum\limits_{k=1}^{N}y_{nk}\hat{y}_{nk}}{\sum\limits_{k=1}^{N}y_{nk}\hat{y}_{nk} + \alpha\sum\limits_{k=1}^{N}y_{nk}\hat{y}_{nk} + \beta\sum\limits_{k=1}^{N}y_{nk}\hat{y}_{nk}}    
\end{equation}

Also by adjusting the hyper-parameters $\alpha$ and $\beta$ (refer to Eq. \ref{eq2}) we can control the trade-off between false positives and false negatives. In our experiments, we set both $\alpha$ and $\beta$ to 0.5. Training with this loss for additional epochs improved the segmentation accuracy on the validation set as well as on the SegTHOR test set, compared to training with the multiclass Dice loss alone.

\subsection{Model Training}
 The adaptive moment estimation (ADAM) optimizer was used to estimate network parameters throughout, and the 1st and 2nd-moment estimates were set to 0.9 and 0.999 respectively. The learning rate was initialized to 0.0001 with a decay factor of 0.2 during training. Validation accuracy was evaluated after each epoch during training, until it stopped increasing. Subsequently, the best performing model was selected for evaluation on the test set. We first trained our model using five-fold cross-validation without any online data augmentation and using only multiclass Dice loss function. In the second stage, in order to improve the segmentation accuracy, we loaded the weights from the first stage and trained the model with random online data augmentation (zooming, rotation, shifting, shearing, and cropping) for 50 additional epochs. This lead to significant performance improvement on the SegTHOR test data. 
 As the multiclass Dice loss does not account for class imbalance, we further improved the second stage of the training process, by employing the TL in place of the former. Consequently, the highest accuracy achieved by our approach employed the TL along with online data augmentation. The network was implemented using Keras, an open-source deep learning library for Python, and was trained on an NVIDIA Titan X-Pascal GPU with 3840 CUDA cores and 12GB RAM. On the test dataset, we observed that our model predicted small structures in implausible locations. 
This was addressed by post-processing the segmentations, to retain only the largest connected component for each structure. As the segmentations predicted by our network were already of good quality, this only lead to marginal improvements in the average Dice score, of approximately 0.002. However, it substantially reduced the average Hausdorff distance, which is very sensitive to outliers.


\subsection{Evaluation Metrics}
\label{sec:pagestyle}
Two standard evaluation metrics are used assess segmentation accuracy, namely, the Dice score coefficient (DSC) and Hausdorff distance (HD). The DSC metric, also known as F1-score, measures the similarity/overlap between manual and automatic segmentation. DSC metric is the most widely used metric to evaluate segmentation accuracy, and is defined as:

\begin{equation}
    DSC (G, P) = \frac{2 TP}{(FP + 2TP + FN)} = \frac{2 |G_{i} \cap P_{i}|}{|G_{i}| + |P_{i}|}
\end{equation}

The HD is defined as the largest of the pairwise distances from points in one set to their corresponding closest points in another set. This is expressed as:
\begin{equation}
    HD (G, P) = \max_{g\in G}\Bigg\{\max_{p\in P}\Bigg\{\sqrt{g^{2}-p^{2}}\Bigg\}\Bigg\}
\end{equation}
In Eq. (3) and (4), $(G)$ and $(P)$ represent the ground truth and predicted segmentations, respectively.

\section{Results and Discussions}
The average DSC and HD measures achieved by 2D U-Net+DR across five-fold cross-validation experiments are summarized in Table \ref{tab:tab1}. The DSC scores achieved by the 2D U-Net+DR without data augmentation for the validation and test sets were 93.61\% and 88.69\%, respectively. The same network with online data augmentation significantly improved the segmentation accuracy to 94.53\% and 91.43\% for the validation and test sets, respectively. Finally, on fine-tuning the trained network using the TL we achieved DSC scores of 94.59\% and 91.57\%, respectively. Compared to \cite{7950685}, our method achieved DSC and HD scores of 85.67\% and 0.30mm for the esophagus, the most difficult OAR to segment. Table \ref{my-label}. illustrates the DSC and HD scores of each individual organ for 2D U-Net+DR method with online augmentation and TL on test data set.

The images presented in Fig.\ref{fig:UNet} help visually assess the segmentation quality of the proposed method on three test volumes. Here, the green color represents the heart, and the red, blue and yellow colors represent the esophagus, trachea, and aorta respectively. 
We obtained the highest average DSC value and HD for the heart and Aorta because of its high contrast, regular shape, and larger size compared to the other organs. As expected, the esophagus had the lowest average DSC and HD values due to its irregularity and low contrast, making it difficult to identify within CT volumes. However, our method achieved a DSC score of 85.8\% for the esophagus on test data set, demonstrating better performance in comparison to the method proposed in \cite{7950685} which used a shape mask network architecture and conditional random fields. These results highlight the effectiveness of the proposed approach for segmenting OARs, which is essential for radiation therapy planning.

\begin{table}[t!]
\resizebox{\columnwidth}{!}{
\begin{tabular}{|l|c|c|c|c|}
\hline
{\textbf{Methods}}                                        & \textbf{Train Data}   & \textbf{Validation Data} & \multicolumn{2}{c|}{\textbf{Test Data}}      \\ \cline{2-5} 
                                                                         & \textbf{DSC {[}\%{]}} & \textbf{DSC {[}\%{]}}    & \textbf{DSC {[}\%{]}} & \textbf{HD {[}mm{]}} \\ \hline
2D U-Net + DR                                                            & 0.9784                & 0.9361                   & 0.8869                & 0.4461               \\ \hline
\begin{tabular}[c]{@{}l@{}}2D U-Net + DR\\ (Augmented)\end{tabular}      & 0.9741                & 0.9453                   & 0.9143                & 0.2536               \\ \hline
\begin{tabular}[c]{@{}l@{}}2D U-Net + DR\\ (Augmented) + TL\end{tabular} & \textbf{0.9749 }               & \textbf{0.9459}                   & \textbf{0.9157}               & \textbf{0.2500}               \\ \hline
\end{tabular}}
\caption{The DSC and HD scores for training, validation and test dataset.}
\label{tab:tab1}
\end{table}

\begin{table}[ht]
\small
\centering
\begin{tabular}{|l|c|c|c|c|}
\hline
\textbf{Metrics}      & \textbf{Esophagus} & \textbf{Heart} & \textbf{Trachea} & \multicolumn{1}{l|}{\textbf{Aorta}} \\ \hline
\textbf{DSC {[}\%{]}} & 0.858              & 0.941          & 0.926            & 0.938                               \\ \hline
\textbf{HD {[}mm{]}}  & 0.331              & 0.226          & 0.193            & 0.297                               \\ \hline
\end{tabular}
\caption{The DSC and HD scores of each organ for 2D U-Net + DR(Augmented) + TL method.}
\label{my-label}
\end{table}

\begin{figure}[htb]
  \centering
  \centerline{\includegraphics[width = 8.6cm]{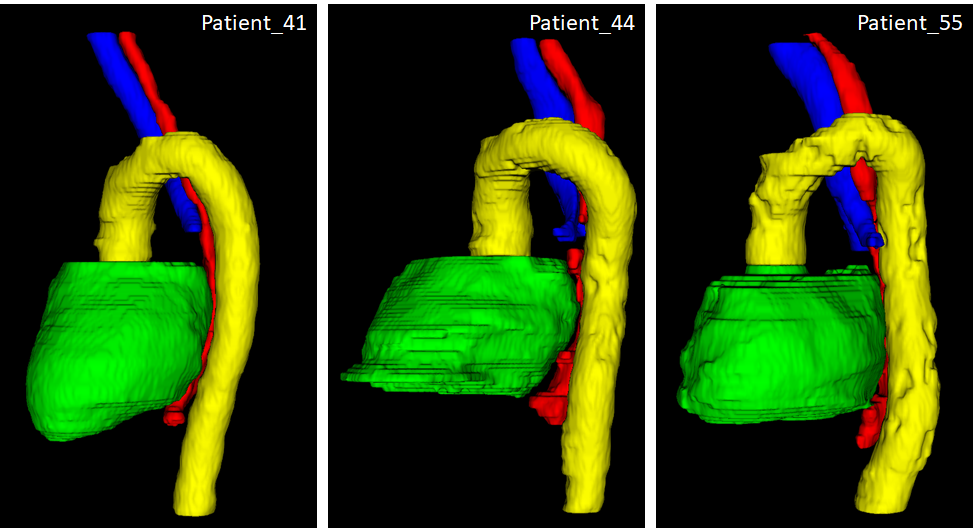}}
\caption{3D surface segmentation outputs of proposed model for three subjects from ISBI SegTHOR challenge test set.}
\label{fig:UNet}
\end{figure}

\section{Conclusions}
In this study, we presented a fully automated approach, called 2D U-Net+DR, for automatic segmentation of the OARs (esophagus, heart, aorta, and trachea) in CT volumes. Our approach provides accurate and reproducible segmentations, thereby aiding in improving consistency and robustness in delineating OARs, relative to manual segmentations. The method uses both local and global information, by expanding the receptive-field in the lowest level of the network, using dilated convolutions. The two-stage training strategy employed here, together with the multi-class soft Dice loss and Tversky loss, significantly improved the segmentation accuracy. Furthermore, we believe that including additional information, e.g. MR images, may be beneficial for some OARs with poorly-visible boundaries such as the esophagus.

\bibliographystyle{IEEEbib}
\bibliography{refs}

\end{document}